\documentclass[10pt]{article} 

\usepackage[accepted]{rlj} 

\usepackage{amssymb}            
\usepackage{mathtools}          
\usepackage{mathrsfs}           
\usepackage{graphicx}           
\usepackage{subcaption}         
\usepackage[space]{grffile}     
\usepackage{url}                
\usepackage{fancyhdr}           
\usepackage{booktabs}           
\usepackage{float}              
\usepackage{algorithm}
\usepackage{algpseudocode}

\usepackage{glossaries}
\glsdisablehyper
\newacronym{ppo}{PPO}{Proximal Policy Optimization}
\newacronym{rtrl}{RTRL}{Real-Time Recurrent Learning}
\newacronym{rtu}{RTU}{Recurrent Trace Units}
\newacronym{gru}{GRU}{Gated Recurrent Unit}
\newacronym{lstm}{LSTM}{Long Short-Term Memory}

\title{Streaming Reinforcement Learning under Partial Observability with Real-Time Recurrent Learning}

\setrunningtitle{Partially Observable Streaming Reinforcement Learning}

\author{Noah Farr\textsuperscript{1,5,$\ast$}, Aryaman Reddi\textsuperscript{1,3,$\ast$}, Carlo D’Eramo\textsuperscript{2}, Jan Peters\textsuperscript{1,3,4}}

\emails{noah.farr@stud.tu-darmstadt.de\, aryaman.reddi@tu-darmstadt.de}

\affiliations{
$^{1}$\textbf{Technical University of Darmstadt}\\
$^{2}$\textbf{Center for Artificial Intelligence and Data Science, University of Würzburg}\\
$^{3}$\textbf{Hessian.AI}\\
$^{4}$\textbf{German Research Center for AI (DFKI)}\\
$^{5}$\textbf{Zuse School ELIZA}\\
\par
$^\ast$ Equal contribution.
}

\begin{document}

\maketitle  

\begin{abstract}
Streaming reinforcement learning has emerged as an online learning paradigm that conforms to the restrictions of natural learning agents that process data incrementally, i.e. with a batch size of 1 and no replay buffer. While streaming RL has recently been shown to scale with deep function approximation with full observability, partially observable settings have remained out of reach.
Truncated backpropagation through time collapses to a one-step gradient
horizon under the streaming setting, and exact real-time recurrent learning is
prohibitively expensive. We close this gap using
recurrent trace units, a diagonal recurrent architecture that enables exact RTRL with linear time and memory complexity in the parameter count, and show that they integrate cleanly into existing streaming algorithms across both discrete and continuous control. On a MemoryChain diagnostic with chain lengths from 2 to 128, our method sustains performance where streaming TBPTT(1) baselines using feedforward, GRU, and RTU networks collapse. On five POPGym tasks and on partially
observable MuJoCo continuous control, the streaming approach is competitive with batched PPO on POPGym and recovers a substantial fraction of batched performance on masked MuJoCo, despite using no replay buffer or batched updates.

\end{abstract}

\section{Introduction}
\label{sec:introduction}

Streaming reinforcement learning, where an agent updates from each sample
as it arrives without a replay buffer, has recently been shown to scale to
deep function approximation in fully observable environments using eligibility traces~\citep{elsayed2024streaming, vasan2024deep}. Extending streaming methods to partially observable settings, which are more representative of realistic deployment scenarios, has remained out of reach. Under single-sample update settings, the only feasible form of recurrent learning with backpropagation through
time is with one-step truncation, denoted TBPTT(1), which provides only a
one-step gradient horizon.

\gls{rtrl}~\citep{williams1989learning} maintains
recurrent forward-mode gradients incrementally and is the natural fit for
streaming, but for general recurrent architectures it has prohibitive
computational cost that scales quartically in the hidden dimension.
\gls{rtu}~\citep{elelimy2024real}, a recently proposed
diagonal complex-valued recurrent architecture, admit exact \gls{rtrl} with linear time and memory complexity in the parameter count, but have so far only been used
with batched algorithms such as \gls{ppo}~\citep{schulman2017proximal}. Bringing these two threads together, we show that recurrent \gls{rtrl} traces with \glspl{rtu} compose naturally with the eligibility traces of the streaming update, yielding a single-pass procedure that propagates credit through both time and the recurrent state without truncation.

\begin{figure}
    \centering
    \begin{subfigure}[t]{0.49\linewidth}
        \centering
        \includegraphics[width=\linewidth]{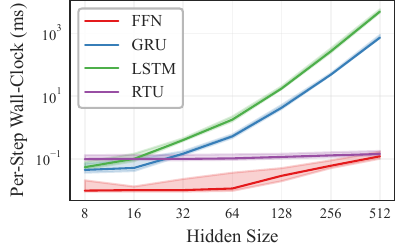}
        \caption{Per-step wall-clock time (ms) of a single gradient update as a function of hidden width, comparing feedforward backpropagation (FFN) against exact \gls{rtrl} for \gls{gru}, LSTM, and \gls{rtu} recurrent layers.}
        \label{fig:rtrl}
    \end{subfigure}
    \hfill
    \begin{subfigure}[t]{0.49\linewidth}
        \centering
        \includegraphics[width=\linewidth]{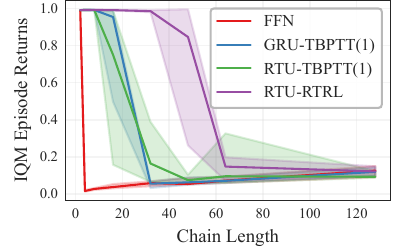}
        \caption{Episodic return as a function of chain length on the MemoryChain diagnostic, averaged over 5 seeds. We compare a feedforward network, GRU and RTU with TBPTT over 1 step and RTU with RTRL.}
        \label{fig:memory_chain}
    \end{subfigure}
\end{figure}

We evaluate the approach on four benchmarks chosen to isolate distinct
properties. A MemoryChain diagnostic with chain lengths from 2 to 128
isolates the challenge of credit assignment over time, where baselines using feedforward, \gls{gru}~\citep{cho2014properties}, and \gls{rtu} architectures all collapse past 16 steps while streaming \gls{rtrl} with \gls{rtu}s sustains performance until 64 steps (Figure~\ref{fig:memory_chain}). 
On five tasks from POPGym \citep{morad2023popgym}, the streaming methods match batched PPO, with the two streaming variants jointly solving all five. On partially observable MuJoCo with masked positions or velocities \citep{ni2022recurrent}, the streaming methods extend to continuous control but trail batched PPO, recovering a substantial fraction of its performance. On a KMemoryChain task we measure and quantify the staleness of \gls{rtrl} sensitivities.
We do not claim state-of-the-art performance on any
single benchmark, as our contribution is closing the partial observability gap in streaming deep RL with a general method.

\paragraph{Contributions.}
\begin{itemize}
    \item A streaming deep RL algorithm for partially observable
    environments using \gls{rtu}s with \gls{rtrl}, requiring no replay buffer or
    batched updates.
    \item A MemoryChain length sweep showing that this combination sustains
    credit assignment past streaming TBPTT(1) baselines using feedforward,
    \gls{gru}, and \gls{rtu} recurrence.
    \item Empirical evidence on five POPGym memory tasks and partially
    observable MuJoCo continuous control that the streaming approach is
    competitive with batched \gls{ppo}.
    \item A staleness analysis of the RTRL sensitivity matrix under streaming updates, bounding the error in steady state and identifying a first-order Taylor correction that tightens the bound.
\end{itemize}

\section{Background}
\label{sec:background}

\paragraph{Streaming reinforcement learning with eligibility traces.}
A streaming agent receives an observation and reward at each time step,
takes an action, performs a learning update from that single sample, and
then discards it. No replay buffer is maintained. To propagate credit
across time without storing past data, streaming algorithms maintain
eligibility traces \citep{sutton1998introduction}. For a per-step gradient
$\boldsymbol{g}_t$ of a value function or policy with respect to parameters $\boldsymbol{w}_t$ at time $t$, these
methods iterate as follows:
\begin{align}
    \boldsymbol{z}_0 & \doteq \boldsymbol{0} \\
    \boldsymbol{z}_t     &= \gamma\lambda \boldsymbol{z}_{t-1} + \boldsymbol{g}_t, \\
    \boldsymbol{w}_{t+1} &= \boldsymbol{w}_t + \alpha \delta_t \boldsymbol{z}_t,
\end{align}
where $\delta_t$ is the temporal difference (TD) error, $\alpha$ is the step size, and
$\lambda \in [0,1]$ is the trace-decay parameter. The total parameter
change generated by these updates is equivalent to that of the
$\lambda$-return in the offline case, without storing a multi-step rollout. Existing streaming
deep RL methods \citep{elsayed2024streaming, vasan2024deep} use this
method with feedforward function approximation, where $\boldsymbol{g}_t$ is the output of standard backpropagation through a single sample.

\paragraph{Real-time recurrent learning.}
For a recurrent network $\boldsymbol{h}_t = \boldsymbol{f}(\boldsymbol{h}_{t-1}, \boldsymbol{x}_t; \boldsymbol{\psi})$ with parameters
$\boldsymbol{\psi}$, computing $\boldsymbol{g}_t$ requires the gradient
$\partial \boldsymbol{h}_t / \partial \boldsymbol{\psi}$ of the current hidden state with respect to
the recurrent parameters. Backpropagation through time (BPTT) obtains this gradient by unrolling the recurrence backward over a stored trajectory of length $T$, which reduces to a one-step gradient horizon when $T=1$. \gls{rtrl}
\citep{williams1989learning} avoids this unrolling by maintaining the gradient forward through time, applying the chain rule to the recurrence,
\begin{equation}
    \frac{\partial \boldsymbol{h}_t}{\partial \boldsymbol{\psi}}
    = \frac{\partial \boldsymbol{f}(\boldsymbol{h}_{t-1},\boldsymbol{x}_t; \boldsymbol{\psi})}{\partial \boldsymbol{\psi}}
    + \frac{\partial \boldsymbol{f}(\boldsymbol{h}_{t-1},\boldsymbol{x}_t; \boldsymbol{\psi})}{\partial \boldsymbol{h}_{t-1}} \,
      \frac{\partial \boldsymbol{h}_{t-1}}{\partial \boldsymbol{\psi}}.
\end{equation}
The first term is the direct dependence of $\boldsymbol{h}_t$ on $\boldsymbol{\psi}$ through the
current step. The second propagates accumulated dependence through
$\boldsymbol{h}_{t-1}$. For a general recurrent network with hidden dimension $n$ and
parameter count $O(n^2)$, this gradient is an $n \times O(n^2)$ matrix
and the recursion costs $O(n^4)$ per step, prohibitive even for moderate
$n$.

\paragraph{Recurrent trace units.}
\gls{rtu}s \citep{elelimy2024real} sidestep this barrier
with a diagonal complex-valued recurrence,
\begin{equation}
    \boldsymbol{h}_t = \boldsymbol{\Lambda} \boldsymbol{h}_{t-1} + \boldsymbol{W} \boldsymbol{x}_t,
\end{equation}
where $\boldsymbol{\Lambda} \in \mathbb{C}^{n\times n}$ is diagonal with entries
$\lambda_k = r_k(\cos\Omega_k + i\sin\Omega_k)$, where $r_k$ and $\Omega_k$ define the polar complex representation of the $k$'th parameter. Because each hidden unit
has only one recurrent connection to itself, exact \gls{rtrl} has linear time and memory complexity in the number of
parameters. Figure~\ref{fig:rtrl} confirms this empirically. Per-step wall-clock time for RTU under exact RTRL tracks feedforward backpropagation as hidden width grows, while GRU and LSTM under RTRL exhibit the prohibitive $O(n^4)$ scaling.

\section{Methodology}
\label{sec:methodology}

Our method makes a single change to existing streaming algorithms.
Where an algorithm typically uses a feedforward function approximator
mapping observations to values or actions, we insert an \gls{rtu} layer between the observation and the feedforward head. The streaming update machinery, including eligibility traces and step-size adaptation, is unchanged. The only modification is how the per-step gradient that feeds into the eligibility
trace is computed. We describe this gradient computation for a single
network below.

\paragraph{Per-step gradient.}
The function approximator has three parameter groups: an encoder
$\phi(\,\cdot\,;\boldsymbol{\theta})$ producing features $\boldsymbol{x}_t$, a recurrent \gls{rtu}
layer with parameters $\boldsymbol{\psi}$ producing the state
$\boldsymbol{h}_t$, and a feedforward head with parameters $\boldsymbol{w}$
(e.g.\ a value head $v(\boldsymbol{h}_t; \boldsymbol{w}_v)$). We compute the gradient
of the head output with respect to each group in a single backward pass,
combined with the \gls{rtu}'s forward-mode \gls{rtrl} trace
$\boldsymbol{S}_t = \partial \boldsymbol{h}_t / \partial \boldsymbol{\psi}$.

The head parameters have no temporal dependency and receive an ordinary
backward pass. The recurrent parameters enter only through
$\boldsymbol{h}_t$, so combining the backward pass through the head with
$\boldsymbol{S}_t$ yields their gradient across the entire history at
$\mathcal{O}(|\boldsymbol{\psi}|)$ per step. The encoder parameters are
more complicated. They influence the state both instantaneously through
$\boldsymbol{x}_t$ and recursively through every past state. Maintaining the exact
trace would cost $\mathcal{O}(|\boldsymbol{\theta}|\cdot\dim \boldsymbol{h})$ per step,
prohibitive for a deep encoder. We therefore use a one-step
approximation, backpropagating through the head, a single step of the
recurrence, and the encoder. This credits the encoder for its immediate
effect on $\boldsymbol{h}_t$ while ignoring its influence through past
states.

The same decomposition applies to any other head (policy, auxiliary).
Each network carries its own \gls{rtu} layer and \gls{rtrl} trace, so
recurrent parameters are updated independently across heads.

\paragraph{Composition with eligibility traces.}
The \gls{rtrl} trace and the eligibility trace coexist without interacting. The
\gls{rtrl} trace is a forward-mode object that maintains
$\partial \boldsymbol{h}_t / \partial \boldsymbol{\psi}$ at each step, with no notion of credit
assignment over time. The eligibility trace is a temporal accumulator of
whatever per-step gradient the streaming algorithm uses, and accumulates
the recurrent contribution as it accumulates the head
contribution. Operationally, each \gls{rtu} layer supplies its
$\partial \boldsymbol{h}_t / \partial \boldsymbol{\psi}$ on demand whenever the streaming update
needs the gradient with respect to its recurrent parameters, and the
eligibility trace then folds this into its usual recursion.

\paragraph{Instantiations.}
We evaluate this method on two streaming algorithms covering both
value-based and policy-based control. We use QRC($\lambda$)
\citep{elelimy2025deep} for value-based control and stream AC($\lambda$)
\citep{elsayed2024streaming} for policy-based control, applying QRC($\lambda$) to MemoryChain, QRC($\lambda$) and stream AC($\lambda$) to POPGym as well as KMemoryChain, and stream AC($\lambda$) to masked MuJoCo. The method itself is
independent of this choice and applies to any streaming algorithm whose
update is driven by per-step gradients. In both cases each network has
its own \gls{rtu} layer with its own \gls{rtrl} trace, so the recurrent parameters
of each network are updated independently. Complete pseudocode for both
instantiations is given in Appendix \ref{sec:pseudocode}.

\section{Experiments}
\label{sec:experiments}
 
We evaluate the method on four benchmarks chosen to isolate distinct properties.
MemoryChain (Section~\ref{sec:memory_chain}) sweeps a memory horizon to measure how far credit propagates through a streaming update.
POPGym (Section~\ref{sec:popgym}) tests whether the streaming method is competitive with a batched recurrent baseline on long-memory discrete control.
Masked MuJoCo (Section~\ref{sec:mujoco}) extends the evaluation to continuous control under partial observability.
KMemoryChain (Section~\ref{sec:staleness}) sweeps a memory length to measure the staleness of sensitivities in \gls{rtrl}.
All curves report the interquartile mean (IQM) over 5 seeds with shaded standard error. All experiments are implemented in Memorax \citep{memorax2025github}, an open-source framework for memory-augmented RL.
Hyperparameters, network architectures, and per-task details are listed in Appendix \ref{sec:hyperparameters}.
 
\subsection{MemoryChain}
\label{sec:memory_chain}
 
\paragraph{Task and design.}
The MemoryChain~\citep{osband2020bsuite} task presents an informative binary cue at $t{=}0$.
The agent then receives uninformative observations for $L{-}1$ steps and is rewarded for reproducing the cue at $t{=}L$.
The chain length $L$ controls how far back the agent must propagate credit, isolating the temporal credit-assignment problem from representation learning.
We sweep $L \in \{2, 4, 8, 16, 32, 48, 64, 128\}$.
The streaming algorithm is QRC$(\lambda)$~\citep{elelimy2025deep}, and we compare four agents.
The first is a feedforward network without recurrence (FFN).
The second is \gls{gru} with TBPTT(1).
The third is \gls{rtu} with TBPTT(1).
The fourth is \gls{rtu} with exact \gls{rtrl}.
All non-recurrence hyperparameters are matched across agents.
 
\paragraph{Result.}
Figure~\ref{fig:memory_chain} reports IQM episodic return against $L$.
The three TBPTT(1) curves drop to near-zero return past $L{=}16$ and stay there for the remainder of the sweep.
A one-step gradient horizon struggles to carry credit through even moderately long chains, and substituting \gls{rtu} for \gls{gru} under TBPTT(1) does not change this.
The \gls{rtu}-\gls{rtrl} curve holds near the maximum return through $L{=}48$ and degrades through $L{=}64$.
The gap between the dotted and solid \gls{rtu} curves attributes this improvement to the gradient method rather than the architecture, since both use the same recurrence and the same streaming update with only the computation of $\partial \boldsymbol{h}_t / \partial \boldsymbol{\psi}$ differing.
\gls{rtrl} is therefore the operative ingredient under streaming, and a diagonal recurrence alone is not sufficient.
 
\subsection{POPGym}
\label{sec:popgym}
 
\paragraph{Task and design.}
We evaluate on five memory tasks from POPGym~\citep{morad2023popgym}.
These are Autoencode, Concentration, CountRecall, HigherLower, and RepeatFirst.
We compare stream AC($\lambda$)-RTU and QRC($\lambda$)-RTU against a batched PPO-RTU baseline and an FFN baseline.
The cell of interest is \gls{rtu} with QRC($\lambda$) and stream AC($\lambda$).
\gls{rtu} with batched \gls{ppo} is the comparable baseline that uses the same architecture without the streaming constraint.
Each agent is run for 5M environment frames.

\begin{figure}[H]
    \centering
    \includegraphics[width=\linewidth]{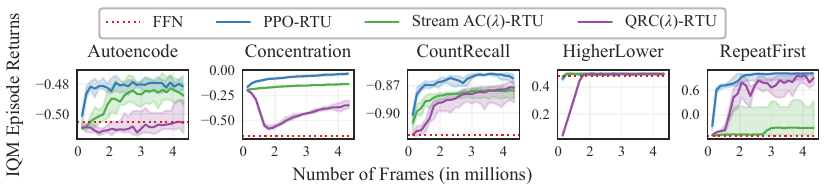}
    \caption{IQM episodic return over 5 seeds with shaded standard error on five POPGym memory tasks.}
    \label{fig:popgym}
\end{figure}
 
\paragraph{Result.}
Across five POPGym tasks, QRC($\lambda$)-\gls{rtu} and stream AC($\lambda$)-\gls{rtu} each outperform the FFN baseline on four of five tasks, isolating the contribution of recurrent state under the streaming constraint. The two methods are complementary across the task set: stream AC($\lambda$)-\gls{rtu} wins on Autoencode, Concentration, and CountRecall, while QRC($\lambda$)-\gls{rtu} wins by a wide margin on RepeatFirst, the one task where stream AC($\lambda$)-\gls{rtu} fails to learn. Compared against batched \gls{ppo}-\gls{rtu}, which relaxes the streaming constraint by maintaining a replay buffer, QRC($\lambda$)-\gls{rtu} matches \gls{ppo} on HigherLower and RepeatFirst, trails it by a small margin on CountRecall, and fails to learn on Autoencode and Concentration—precisely the tasks where stream AC($\lambda$)-\gls{rtu} succeeds. Taken together, the two streaming methods solve all five tasks, recovering batched performance without storing past trajectories.

\subsection{Masked MuJoCo}
\label{sec:mujoco}
 
\paragraph{Task and design.}
Following~\citet{ni2022recurrent}, we evaluate on four MuJoCo locomotion tasks of Ant, HalfCheetah, Hopper, Walker2d and under two observation regimes masked velocities (\textbf{P}) and masked positions (\textbf{V}).
Both settings have partially-observable observations, requiring the agent to infer the missing component from history.
We compare a stream AC($\lambda$) agent with \gls{rtu} against PPO with \gls{rtu}.
Each run uses 2M environment frames.

\begin{figure}[h]
    \centering
    \includegraphics[width=\linewidth]{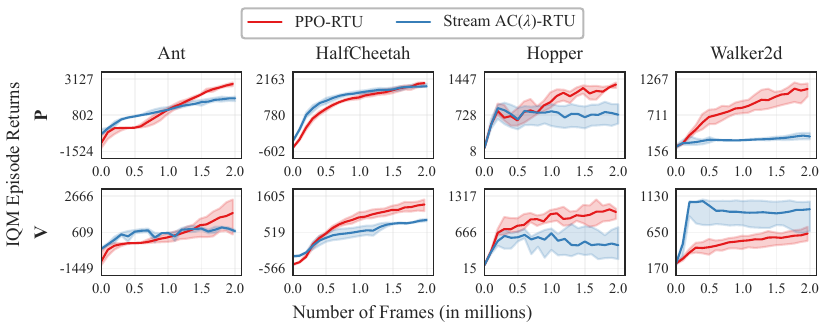}
    \caption{IQM episodic return over 5 seeds with shaded standard error on four MuJoCo tasks under two observation regimes.}
    \label{fig:brax}
\end{figure}

\paragraph{Result.}
Across the eight task-regime cells, stream AC($\lambda$)-\gls{rtu} matches
batched PPO-\gls{rtu} on HalfCheetah-P, overtakes it on Walker2d-V, fails to
learn on Walker2d-P, and trails it by varying margins on the remaining five.
HalfCheetah-P is essentially tied, both curves rise together and finish
indistinguishably near $2{,}000$. Walker2d-V is the reversal, where stream AC($\lambda$)
rises sharply within the first $0.5$M frames to roughly $800$ and holds that
level while PPO climbs more slowly and finishes below it. Walker2d-P is the only failure case where stream AC($\lambda$) remains near its initial return throughout
training while PPO reaches roughly $1{,}000$. On the remaining five cells both
methods learn, with stream AC($\lambda$) trailing PPO by a modest margin on Hopper in
both regimes, and by a wider margin on Ant in both regimes and on
HalfCheetah-V, where stream AC($\lambda$) plateaus while PPO continues to improve.
Partially observable continuous control is therefore feasible under the
streaming constraint, but the gap to batched PPO is wider than on POPGym,
indicating that the streaming method recovers a meaningful fraction, though
not all, of batched performance in this regime.

\subsection{RTRL Staleness}
\label{sec:staleness}

The RTRL sensitivity matrix $\boldsymbol{S}_t = \partial \boldsymbol{h}_t / \partial \boldsymbol{\psi}$ is built up by accumulating immediate Jacobians $\boldsymbol{I}_k(\boldsymbol{\psi}_k)$ injected at past steps and propagated forward by the state-to-state Jacobian $\boldsymbol{J}_t$. Under a streaming update the parameters $\boldsymbol{\psi}$ move every step, so each $\boldsymbol{I}_k(\boldsymbol{\psi}_k)$ in this sum was evaluated at parameters the optimizer has since left behind. The maintained sensitivity is therefore a biased estimate of the ideal sensitivity $\boldsymbol{S}_t^\ast$, which re-evaluates every historical immediate Jacobian at the current parameters $\boldsymbol{\psi}_t$. 

RTRL staleness has been observed qualitatively in prior work \citep{williams1989learning,menick2021practical,irie2024exploring,elelimy2024real}, with smaller learning rates \citep{menick2021practical} or periodic recomputation of the trace over the trajectory \citep{elelimy2024real} suggested as mitigations, but it has not been quantitatively bounded.

Appendix~\ref{sec:staleness-proof} bounds this staleness in steady state and shows that a first-order Taylor correction in $\boldsymbol{\psi}$ reduces the error from $\mathcal{O}(\eta)$ to $\mathcal{O}(\eta^2)$. This section probes the bias empirically and tests whether the correction tracks the bound under sustained credit-assignment load.

\paragraph{Task and design.}
We use KMemoryChain, an every-step variant of MemoryChain designed to exercise the RTRL trace continuously. At each step the agent observes a fresh $\pm 1$ bit together with a time-remaining feature, and is rewarded $\pm 1$ for predicting the bit it saw $K$ steps ago. Rewards activate after a $K$-step warmup. The classic MemoryChain rewards a single recall at episode end and leaves the immediate Jacobian inactive for most of the rollout, which is precisely the regime in which staleness has the least opportunity to accumulate. KMemoryChain instead injects a new bit and emits a reward-relevant target at every step, keeping the immediate Jacobian non-zero throughout the rollout and making $K$ a direct dial on the credit-assignment horizon over which the sensitivity must remain accurate. We sweep $K \in \{4, 8, 16\}$ and run both streaming algorithms, QRC($\lambda$) and stream AC($\lambda$). Each algorithm is run with the standard RTRL sensitivity and with the first-order Taylor correction of Appendix~\ref{sec:staleness-proof} enabled, giving four configurations per $K$.

\begin{figure}
    \centering
    \begin{subfigure}[t]{0.49\linewidth}
        \centering
        \includegraphics[width=\linewidth]{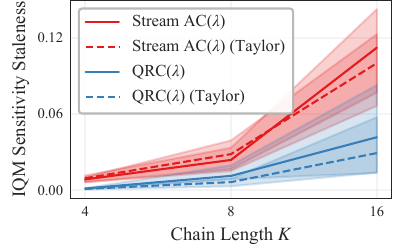}
        \caption{RTRL sensitivity staleness on KMemoryChain. Relative $\ell_2$ distance between the online sensitivity matrix $\tilde{\boldsymbol{S}}_t$ and its recomputed reference $\boldsymbol{S}_t^\ast$, $\lVert \tilde{\boldsymbol{S}}_t - \boldsymbol{S}_t^\ast \rVert_2 / \lVert \boldsymbol{S}_t^\ast \rVert_2$,
over different chain lengths.}
        \label{fig:staleness}
    \end{subfigure}
    \hfill
    \begin{subfigure}[t]{0.49\linewidth}
        \centering
        \includegraphics[width=\linewidth]{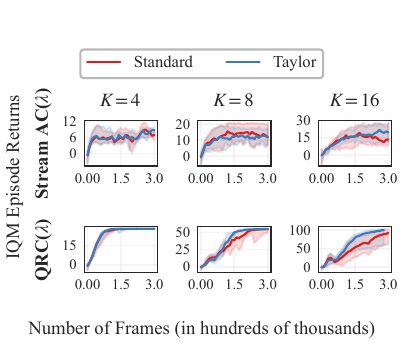}
        \caption{IQM episodic return over 5 seeds with shaded standard error on different chain lengths of KMemoryChain.}
        \label{fig:stale_returns}
    \end{subfigure}
\end{figure}

\paragraph{Staleness metric.}
At every training step we maintain a buffer of the observations from the current episode. To form a reference, we replay the buffer forward from the saved initial state using the \emph{current} parameters $\boldsymbol{\psi}_t$, recomputing what the sensitivity would be if every immediate Jacobian in the episode had been evaluated at $\boldsymbol{\psi}_t$ rather than at the parameters in force when it was originally injected. The staleness reported is the $\ell_2$ distance between the live sensitivity carried by the algorithm and this recomputed reference. 

\paragraph{Result.}
Figure~\ref{fig:staleness} reports the IQM sensitivity staleness on KMemoryChain
as a function of the memory credit-assignment horizon $K$. Staleness rises
with $K$ across all four configurations, remaining near zero at $K{=}4$ and
growing through $K{=}8$ to its peak at $K{=}16$. This $K$-dependence is consistent
with the $(1-\rho)^{-2}$ scaling of the steady-state bound in
Equation~\eqref{eq:err_bound}: longer credit-assignment horizons require the
recurrence to retain memory over more steps, raising the effective contraction
constant and amplifying the bound.

The two algorithms differ in absolute staleness level: stream AC($\lambda$)
staleness is consistently higher than QRC($\lambda$) staleness and grows more
steeply with $K$. We attribute this gap to the optimizer rather than the
algorithm. Stream AC($\lambda$) uses ObGD~\citep{elsayed2024streaming} with
base step size $\alpha=1$, which is scaled down only when the
effective-step-size threshold is violated and otherwise drives large per-step
parameter updates. QRC($\lambda$), by contrast, uses plain SGD with fixed
learning rates of $10^{-4}$ for the Q-network and $10^{-5}$ for the auxiliary
$h$-network, two to three orders of magnitude smaller than the nominal ObGD
step. Because the steady-state bound in Equation~\eqref{eq:err_bound} scales
linearly with the per-step parameter update magnitude $\eta G$, the larger
ObGD updates translate directly into proportionally larger steady-state
staleness. The Taylor correction reduces staleness across both algorithms,
with the largest absolute reduction at $K{=}16$ where the uncorrected
staleness is largest and the leading-order term predicted by
Equation~\eqref{eq:err_recur} dominates. At $K{=}4$ the corrected and
uncorrected variants are essentially indistinguishable, leaving little
headroom for a first-order correction to act on.

Figure~\ref{fig:stale_returns} reports the corresponding IQM episodic returns
across training for each $(\text{algorithm}, K)$ cell. The Taylor correction
does not degrade learning at any $K$, and at $K{=}4$ and $K{=}8$ it tracks the
standard sensitivity throughout training. At $K{=}16$, where the standard
sensitivity incurs the largest staleness, the Taylor correction appears to
learn faster and produce a more stable training curve, providing downstream
evidence that the staleness reductions in Figure~\ref{fig:staleness} translate
into improved credit assignment in the regime where the bound is tightest.
Taken together, streaming RTRL staleness stays bounded under sustained
credit-assignment load, scales with $K$ in the direction the bound predicts,
and tightens under the Taylor correction in the regime where the leading-order
term dominates.

\section{Related Work}
\paragraph{Streaming deep reinforcement learning.}
Stream-x \citep{elsayed2024streaming} stabilized value-based and
actor-critic streaming methods in deep RL through a combination of
step-size adaptation, layer normalization, and observation and reward
scaling, removing the long-standing argument that streaming sacrifices
sample efficiency. AVG \citep{vasan2024deep} obtained a similar result
for policy gradient methods using a reparameterized policy. Both works
restrict to fully observable environments. Our method extends this
line to partial observability while preserving the no-replay,
constant-memory streaming property that motivates it.

\paragraph{Practical real-time recurrent learning.}
\gls{rtrl} was introduced as the natural online counterpart to backpropagation
through time \citep{williams1989learning}, but its quartic cost in the
hidden dimension has long limited its use. Recent work makes \gls{rtrl}
efficient by restricting the recurrence. 
\citet{menick2021practical} sparsify the influence matrix to make RTRL tractable for general recurrent cells and observe that online weight updates introduce a staleness bias in the maintained sensitivity.
Linear recurrent units
\citep{orvieto2023resurrecting} use a diagonal complex-valued recurrence
trained by parallel scan, and \gls{rtu}s \citep{elelimy2024real} build on this
foundation with a real-valued cosine parameterization that fits cleanly
into deep RL pipelines. 
\citet{irie2024exploring} apply exact RTRL with element-wise recurrent networks to actor-critic RL and discuss staleness as a practical limitation.
\gls{rtu}s have so far been combined only with
batched algorithms such as \gls{ppo}. We use them as the recurrent component
of a streaming algorithm, leveraging the linear-time \gls{rtrl} trace they
admit and quantify the staleness that streaming updates induce in Appendix~\ref{sec:staleness-proof}.

\paragraph{Recurrent RL under partial observability.}
Recurrent model-free RL with \gls{gru} or \gls{lstm}~\citep{hochreiter1997long} backbones trained via truncated
BPTT is a strong baseline for POMDPs when a replay buffer is available
\citep{ni2022recurrent}. POPGym \citep{morad2023popgym} provides
systematic benchmarking across recurrent architectures and supplies the
long-memory tasks we use. Our method matches the structural
assumption of these recurrent baselines (model-free, recurrent state for
memory) but removes the dependence on stored trajectories.

\section{Discussion and Limitations}
\label{sec:discussion}

\paragraph{Restricted recurrent architecture.}
We use a single diagonal \gls{rtu} layer per network, because \gls{rtu}s enable exact \gls{rtrl} in linear time and because
multi-layer \gls{rtrl} remains an open challenge. Even with diagonal recurrence at each
layer, the gradient trace between layers requires either approximation
or persisting cross-layer Jacobians. \citet{elelimy2024real} show that
diagonal recurrence does not hurt on partially observable
tasks, which we also observe. Partially observable tasks requiring richer
representations remain to be
tested.

\paragraph{Scope of the empirical claim.}
We do not claim state-of-the-art absolute performance on any single
benchmark. The contribution is closing the streaming-plus-partial-%
observability gap, and we compare against methods that respect the same
resource constraints. On benchmarks where unconstrained batched methods
are available, those methods may still produce better sample efficiency
per environment step. The case for the streaming method is the
regime where the constraints are real, such as on-device learning,
real-time control, and lifelong learning settings where storing past
trajectories is impractical.

\section{Conclusion}
\label{sec:conclusion}

We present a streaming deep reinforcement learning method for partially observable environments. An \gls{rtu} layer trained by exact real-time recurrent learning is inserted between the observation and the feedforward heads of an existing streaming algorithm, leaving the streaming update machinery unchanged. The \gls{rtrl} trace and the streaming eligibility trace compose without modification, yielding a single-pass procedure that propagates credit through both the recurrent state and time. Empirically, the method sustains credit assignment on MemoryChain chain lengths well past streaming TBPTT(1) baselines, jointly solves all five POPGym memory tasks at parity with batched PPO across the two streaming variants, and extends to continuous control on masked MuJoCo at a meaningful performance level. An analysis of the RTRL sensitivity matrix bounds its staleness under streaming updates and identifies a first-order Taylor correction that tightens the bound.


\newpage

\appendix
\section{Hyperparameters}
  \label{sec:hyperparameters}

We used fully connected networks of width $64$ for MemoryChain, POPGym and Masked
MuJoCo, with LeakyReLU activations
on the discrete-action benchmarks and $\tanh$ on Masked MuJoCo, and
LayerNorm \citep{ba2016layer} before each activation. A single \gls{rtu}
layer of hidden dimension $192$ is inserted between the input and
output blocks. We used separate networks for the policy and value
functions. For stream AC($\lambda$) we used softmax policy parameterization in
discrete control and a state-independent Gaussian with actions clipped
to $[-1, 1]$ in continuous control. For QRC($\lambda$) we used an
$\varepsilon$-greedy policy. We used sparse
initialization with a sparsity ratio of $90\%$
\citep{elsayed2024streaming} and normalized observations and rewards
online with running statistics. stream AC($\lambda$) uses ObGD with step size
$\alpha = 1$; the per-benchmark $\kappa$ values for the policy and
value networks are listed in the experiment-specific tables below. QRC($\lambda$)
uses SGD with global-norm gradient clipping at $1.0$ and learning rates
$10^{-4}$ for the Q-network and $10^{-5}$ for the auxiliary
$h$-network. For the batched \gls{ppo} baseline we used the
hyperparameters reported by \citet{elelimy2024real}.

\begin{table}[H]
\centering
\caption{Hyperparameters for QRC($\lambda$) on the MemoryChain experiment
(Section~\ref{sec:memory_chain}).}
\label{tab:hparams-MemoryChain}
\begin{tabular}{ll}
\toprule
Name & Value \\
\midrule
Chain length, $L$                      & $\{2, 4, 8, 16, 32, 48, 64, 128\}$ \\
Frames per run                         & $500{,}000$ \\
Number of seeds                        & $5$ \\
Discount factor, $\gamma$              & $0.99$ \\
Trace decay, $\lambda$                 & $0.95$ \\
$Q$-network step size, $\alpha_Q$      & $10^{-4}$ \\
$h$-network step size, $\alpha_h$      & $10^{-5}$ \\
Regularization coefficient, $\beta$ & $1.0$ \\
MLP width                              & $64$ \\
\gls{rtu} hidden dimension             & $192$ \\
\bottomrule
\end{tabular}
\end{table}

\begin{table}[H]
\centering
\caption{Hyperparameters for QRC$(\lambda)$ on POPGym (Section~\ref{sec:popgym}).}
\label{tab:hparams-popgym-streaming}
\begin{tabular}{ll}
\toprule
Name & Value \\
\midrule
Frames per run                         & $5{,}000{,}000$ \\
Number of seeds                        & $5$ \\
Discount factor, $\gamma$              & $0.99$ \\
Trace decay, $\lambda$                 & $0.95$ \\
$Q$-network step size, $\alpha_Q$      & $10^{-4}$ \\
$h$-network step size, $\alpha_h$      & $10^{-5}$ \\
Regularization coefficient, $\beta$ & $1.0$ \\
$\varepsilon$ schedule             & $1.0 \to 0.01$ over $20\%$ of training \\
MLP width                              & $64$ \\
\gls{rtu} hidden dimension             & $192$ \\
\bottomrule
\end{tabular}
\end{table}

\begin{table}[H]
  \centering
  \caption{Hyperparameters for stream AC$(\lambda)$ on POPGym (Section~\ref{sec:popgym}).}
  \label{tab:hparams-popgym-streamac}
  \begin{tabular}{ll}
  \toprule
  Name & Value \\
  \midrule
  Frames per run                         & $5{,}000{,}000$ \\
  Number of seeds                        & $5$ \\
  Discount factor, $\gamma$              & $0.99$ \\
  Trace decay, $\lambda$                 & $0.8$ \\
  Entropy coefficient                    & $0.095$ \\
  Optimizer                              & ObGD \\
  Actor step size, $\alpha_\pi$          & $1.0$ \\
  Critic step size, $\alpha_V$           & $1.0$ \\
  Actor $\kappa_\pi$                     & $3.0$ \\
  Critic $\kappa_V$                      & $2.0$ \\
  MLP width                              & $64$ \\
  \gls{rtu} hidden dimension             & $192$ \\
  \bottomrule
  \end{tabular}
  \end{table}

\begin{table}[H]
\centering
\caption{Hyperparameters for stream AC($\lambda$) on the Masked MuJoCo experiment
(Section~\ref{sec:mujoco}).}
\label{tab:hparams-mujoco}
\begin{tabular}{ll}
\toprule
Name & Value \\
\midrule
Frames per run                         & $2{,}000{,}000$ \\
Number of seeds                        & $5$ \\
Observation regimes                    & masked velocities (\textbf{P}), masked positions (\textbf{V}) \\
Discount factor, $\gamma$              & $0.99$ \\
Trace decay, $\lambda$                 & $0.8$ \\
Entropy coefficient                    & $0.01$ \\
Optimizer                              & ObGD \\
Actor step size, $\alpha_\pi$          & $1.0$ \\
Critic step size, $\alpha_V$           & $1.0$ \\
Actor $\kappa_\pi$                     & $1.5$ \\
Critic $\kappa_V$                      & $1.0$ \\
MLP width                              & $64$ \\
\gls{rtu} hidden dimension             & $192$ \\
\bottomrule
\end{tabular}
\end{table}

  \begin{table}[H]
  \centering
  \caption{Hyperparameters for stream AC($\lambda$) on the KMemoryChain experiment
  (Section~\ref{sec:staleness}).}
  \label{tab:hparams-StreamAC-KMemoryChain}
  \begin{tabular}{ll}
  \toprule
  Name & Value \\
  \midrule
  Memory length, $K$                     & $\{4, 8, 16\}$ \\
  Frames per run                         & $300{,}000$ \\
  Number of seeds                        & $5$ \\
  Discount factor, $\gamma$              & $0.99$ \\
  Trace decay, $\lambda$                 & $0.95$ \\
  Actor step size, $\alpha_\pi$          & $1.0$ \\
  Critic step size, $\alpha_v$           & $1.0$ \\
  Actor ObGD $\kappa_\pi$                & $3$ \\
  Critic ObGD $\kappa_v$                 & $2.0$ \\
  Entropy coefficient                    & $0.01$ \\
  \gls{rtu} feature dimension            & $64$ \\
  \gls{rtu} hidden dimension             & $192$ \\
  \bottomrule
  \end{tabular}
  \end{table}

  \begin{table}[H]
  \centering
  \caption{Hyperparameters for QRC($\lambda$) on the KMemoryChain experiment
  (Section~\ref{sec:staleness}).}
  \label{tab:hparams-QRC-KMemoryChain}
  \begin{tabular}{ll}
  \toprule
  Name & Value \\
  \midrule
  Memory length, $K$                     & $\{4, 8, 16\}$ \\
  Frames per run                         & $300{,}000$ \\
  Number of seeds                        & $5$ \\
  Discount factor, $\gamma$              & $0.99$ \\
  Trace decay, $\lambda$                 & $0.95$ \\
  $Q$-network step size, $\alpha_Q$      & $10^{-4}$ \\
  $h$-network step size, $\alpha_h$      & $10^{-5}$ \\
  Regularization coefficient, $\beta$    & $1.0$ \\
  $\epsilon$-greedy start                & $1.0$ \\
  $\epsilon$-greedy end                  & $0.01$ \\
  $\epsilon$-greedy decay fraction       & $0.1$ \\
  \gls{rtu} feature dimension            & $64$ \\
  \gls{rtu} hidden dimension             & $192$ \\
  \bottomrule
  \end{tabular}
  \end{table}

\section{Pseudocode}
\label{sec:pseudocode}

  The contribution to either parent algorithm is the per-step gradient
  computation: each network maintains an RTU hidden state $\boldsymbol{h}$ and a
  forward-mode RTRL sensitivity matrix $S = \partial \boldsymbol{h} / \partial \boldsymbol{\psi}$,
  both advanced once per environment step, and gradients of any scalar
  depending on $\boldsymbol{h}$ are computed from $S$. Following the
  parent algorithms, $\boldsymbol{\theta}$, $\boldsymbol{\psi}$ and $\boldsymbol{w}$ denote each network's full
  parameter set (encoder, RTU, head). Trace recursions and update rules are
  unchanged from the parents.

  \begin{algorithm}[H]
  \caption{QRC$(\lambda)$ with RTU-RTRL}
  \label{alg:qrc_rtu}
  \begin{algorithmic}
    \State{Input: a differentiable action-value function parametrization $\hat{q}_{\boldsymbol{w}}$ with an RTU layer}
    \State{Input: a differentiable auxiliary function parametrization $\hat{h}_{\boldsymbol{\theta}}$ with an RTU layer}
    \State{Algorithm parameters: learning rate $\alpha_q$, $h$ learning rate $\alpha_h$, exploration parameter $\epsilon$.}
    \State{Initialize $\boldsymbol{z}_t^{\boldsymbol{\theta}} \leftarrow \boldsymbol{0}$}
    \State{Initialize $\boldsymbol{z}_t^{\boldsymbol{w}} \leftarrow \boldsymbol{0}$}
    \State{Initialize $z_t^h \leftarrow 0$}
    \State{Initialize RTU carries $\boldsymbol{h}_t^q, \boldsymbol{h}_t^h \leftarrow \boldsymbol{0}$ and RTRL traces $\boldsymbol{S}_t^q, \boldsymbol{S}_t^h \leftarrow \boldsymbol{0}$}
    \State{Observe initial state $\boldsymbol{S}_0$}
    \For{iteration $t = 1, 2, \cdots$}
      \State{Advance $(\boldsymbol{h}_t^q, \boldsymbol{S}_t^q)$ and $(\boldsymbol{h}_t^h, \boldsymbol{S}_t^h)$ one RTU step on $\boldsymbol{S}_t$.}
      \State{Sample an action $A_t \sim \pi$.} \Comment{We use an $\epsilon$-greedy policy.}
      \State{Take action $A_t$, observe $R_{t+1}$ and $S_{t+1}$.}
      \State{Compute $\delta_t$ and $\nabla_{\boldsymbol{w}_t} \delta_t$ as in QRC$(\lambda)$ \citep{elelimy2025deep}, with parameter gradients of $\hat q_{\boldsymbol{w}}$ obtained from the
  RTRL trace $\boldsymbol{S}_t^q$.}
      \State{Update the traces $\boldsymbol{z}_t^{\boldsymbol{\theta}}$, $\boldsymbol{z}_t^{\boldsymbol{w}}$, and $z_t^h$ as in QRC$(\lambda)$ \citep{elelimy2025deep}, with parameter
  gradients of $\hat h_{\boldsymbol{\theta}}$ obtained from the RTRL trace $\boldsymbol{S}_t^h$.}
      \State{Compute $\Delta \boldsymbol{w}_t$ and $\Delta \boldsymbol{\theta}_t$ as in QRC$(\lambda)$ \citep{elelimy2025deep}.}
      \State{Update the parameters $\boldsymbol{w}_{t+1} \leftarrow \boldsymbol{w}_t + \alpha_q \Delta \boldsymbol{w}_t$.}
      \State{Update the parameters $\boldsymbol{\theta}_{t+1} \leftarrow \boldsymbol{\theta}_t + \alpha_h \Delta \boldsymbol{\theta}_t$.}
      \If{episode terminated or $A_t$ is non-greedy}
        \State{Reset the traces $\boldsymbol{z}_t^{\boldsymbol{\theta}}$, $\boldsymbol{z}_t^{\boldsymbol{w}}$, and $z_t^h$ to zeros.}
      \EndIf
      \If{episode terminated}
        \State{Reset the RTU carries $\boldsymbol{h}_t^q, \boldsymbol{h}_t^h$ and RTRL traces $\boldsymbol{S}_t^q, \boldsymbol{S}_t^h$ to zeros.}
      \EndIf
    \EndFor
  \end{algorithmic}
  \end{algorithm}

 \begin{algorithm}[H]
  \caption{stream AC$(\lambda)$ with RTU-RTRL}
  \label{alg:streamac_rtu}
  \begin{algorithmic}
    \State{Input: a differentiable policy parametrization $\pi(a \mid s; \boldsymbol{\theta})$ with an RTU layer}
    \State{Input: a differentiable state-value function parametrization $\hat{v}(s; \boldsymbol{w})$ with an RTU layer}
    \State{Algorithm parameters: discount $\gamma$, trace decay $\lambda$, policy step size $\alpha_\pi$, value step size $\alpha_{\hat v}$, policy and value scaling factors $\kappa_\pi,
  \kappa_{\hat v}$, entropy coefficient $\tau$.}
    \State{Initialize $\boldsymbol{z}_t^{\boldsymbol{\theta}} \leftarrow \boldsymbol{0}$}
    \State{Initialize $\boldsymbol{z}_t^{\boldsymbol{w}} \leftarrow \boldsymbol{0}$}
    \State{Initialize RTU carries $\boldsymbol{h}_t^\pi, \boldsymbol{h}_t^v \leftarrow \boldsymbol{0}$ and RTRL traces $\boldsymbol{S}_t^\pi, \boldsymbol{S}_t^v \leftarrow \boldsymbol{0}$}
    \State{Initialize observation- and reward-normalization statistics as in stream AC$(\lambda)$ \citep{elsayed2024streaming}}
    \State{Observe initial state $\boldsymbol{S}_0$}
    \For{iteration $t = 1, 2, \cdots$}
      \State{Advance $(\boldsymbol{h}_t^\pi, \boldsymbol{S}_t^\pi)$ and $(\boldsymbol{h}_t^v, \boldsymbol{S}_t^v)$ one RTU step on $\boldsymbol{S}_t$.}
      \State{Sample an action $A_t \sim \pi(\cdot \mid \boldsymbol{S}_t, \boldsymbol{\theta}_t)$.}
      \State{Take action $A_t$, observe $R_{t+1}$ and $S_{t+1}$.}
      \State{Compute $\delta_t$ and the per-step gradients $\nabla_{\boldsymbol{w}} \hat{v}(\boldsymbol{S}_t, \boldsymbol{w}_t)$ and $\nabla_{\boldsymbol{\theta}} \bigl( \log \pi(A_t \mid \boldsymbol{S}_t,
  \boldsymbol{\theta}_t) + \tau\, \mathrm{sign}(\delta_t)\, \mathcal{H}(\cdot \mid \boldsymbol{S}_t, \boldsymbol{\theta}_t) \bigr)$ as in stream AC$(\lambda)$ \citep{elsayed2024streaming}, with parameter
  gradients of $\hat v$ and $\pi$ obtained from the RTRL traces $\boldsymbol{S}_t^v$ and $\boldsymbol{S}_t^\pi$.}
      \State{Update the traces $\boldsymbol{z}_t^{\boldsymbol{\theta}}$ and $\boldsymbol{z}_t^{\boldsymbol{w}}$ as in stream AC$(\lambda)$ \citep{elsayed2024streaming}.}
      \State{Compute $\Delta \boldsymbol{\theta}_t$ and $\Delta \boldsymbol{w}_t$ via ObGD as in stream AC$(\lambda)$ \citep{elsayed2024streaming}.}
      \State{Update the parameters $\boldsymbol{\theta}_{t+1} \leftarrow \boldsymbol{\theta}_t + \Delta \boldsymbol{\theta}_t$.}
      \State{Update the parameters $\boldsymbol{w}_{t+1} \leftarrow \boldsymbol{w}_t + \Delta \boldsymbol{w}_t$.}
      \If{episode terminated}
        \State{Reset the traces $\boldsymbol{z}_t^{\boldsymbol{\theta}}$ and $\boldsymbol{z}_t^{\boldsymbol{w}}$ to zeros.}
        \State{Reset the RTU carries $\boldsymbol{h}_t^\pi, \boldsymbol{h}_t^v$ and RTRL traces $\boldsymbol{S}_t^\pi, \boldsymbol{S}_t^v$ to zeros.}
      \EndIf
    \EndFor
  \end{algorithmic}
  \end{algorithm}

\section{Staleness error bound for the \gls{rtrl} sensitivity matrix}
\label{sec:staleness-proof}

In the nonlinear setting, the \gls{rtrl} sensitivity matrix maintained by an
\gls{rtu} layer incurs a \emph{staleness error}: immediate Jacobians
accumulated at earlier steps were evaluated at parameters that the
optimizer has since moved away from. We bound this error and give a
second-order correction that reduces it from $\mathcal{O}(\eta)$ to
$\mathcal{O}(\eta^2)$.

\subsection{Setup and assumptions}
\label{sec:staleness-setup}

For a recurrent network $\boldsymbol{h}_t = \boldsymbol{f}(\boldsymbol{h}_{t-1}, \boldsymbol{x}_t; \boldsymbol{\psi})$ with state-to-state
Jacobian $\boldsymbol{J}_t = \partial \boldsymbol{h}_t / \partial \boldsymbol{h}_{t-1}$ and immediate Jacobian
$\boldsymbol{I}_t(\boldsymbol{\psi}) = \partial \boldsymbol{h}_t / \partial \boldsymbol{\psi}$, \gls{rtrl} maintains the sensitivity
matrix
\begin{equation}
    \boldsymbol{S}_t \;=\; \boldsymbol{J}_t\, \boldsymbol{S}_{t-1} \;+\; \boldsymbol{I}_t.
\end{equation}
Let $\boldsymbol{\psi}_t \in \mathbb{R}^p$ denote the recurrent parameters in force at
step $t$. The \emph{stale sensitivity matrix} maintained by the online
algorithm is
\begin{equation}
    \tilde{\boldsymbol{S}}_t \;=\; \boldsymbol{J}_t\, \tilde{\boldsymbol{S}}_{t-1} \;+\; \boldsymbol{I}_t(\boldsymbol{\psi}_t)
    \;=\; \sum_{k=0}^{t} \left(\prod_{j=k+1}^{t} \boldsymbol{J}_j\right) \boldsymbol{I}_k(\boldsymbol{\psi}_k),
\end{equation}
where the immediate Jacobian injected at step $k$ was evaluated at the
parameters $\boldsymbol{\psi}_k$ in force at that step. The \emph{ideal sensitivity
matrix}, which re-evaluates each historical immediate Jacobian at the
current parameters, is
\begin{equation}
    \boldsymbol{S}_t^\ast \;=\; \sum_{k=0}^{t} \left(\prod_{j=k+1}^{t} \boldsymbol{J}_j\right) \boldsymbol{I}_k(\boldsymbol{\psi}_t).
\end{equation}
The staleness error is $E_t = \|\boldsymbol{S}_t^\ast - \tilde{\boldsymbol{S}}_t\|$.

\paragraph{Assumptions.}
\begin{itemize}
    \item[(A1)] \textbf{Contraction.} $\|\boldsymbol{J}_t\| \le \rho < 1$ uniformly in
    $t$.\footnote{For an \gls{rtu} layer with
    $\boldsymbol{\Lambda} = \mathrm{diag}(r_k e^{i\theta_k})$ the contraction constant
    is $\rho = \max_k r_k$, kept below one by the standard \gls{rtu}
    parameterization.}
    \item[(A2)] \textbf{Bounded updates.}
    $\|\boldsymbol{\psi}_t - \boldsymbol{\psi}_{t-1}\| \le \eta G$ for all $t$, where $\eta$ is a
    step size and $G$ bounds the per-step update magnitude.
    \item[(A3)] \textbf{Lipschitz immediate Jacobian.} The map
    $\boldsymbol{\psi} \mapsto \boldsymbol{I}_t(\boldsymbol{\psi})$ is Lipschitz with constant $L_I$,
    uniformly in $t$:
    $\|\boldsymbol{I}_t(\boldsymbol{\psi}) - \boldsymbol{I}_t(\boldsymbol{\psi'})\| \le L_I \|\boldsymbol{\psi} - \boldsymbol{\psi'}\|$.
    \item[(A4)] \textbf{Bounded magnitude.}
    $\|\boldsymbol{I}_t(\boldsymbol{\psi})\| \le M_I$ uniformly.
\end{itemize}

\paragraph{Bound on the sensitivity matrix.}
Taking norms in the stale recursion and applying the triangle inequality,
$\|\tilde{\boldsymbol{S}}_t\| \le \rho \|\tilde{\boldsymbol{S}}_{t-1}\| + M_I$,
which unrolls to the geometric steady-state bound
$\|\tilde{\boldsymbol{S}}_t\| \le M_I / (1 - \rho)$.

\paragraph{Error recurrence.}
Expanding $\boldsymbol{S}_{t+1}^\ast$ and $\tilde{\boldsymbol{S}}_{t+1}$, the injection at step $t+1$
cancels because both contain $\boldsymbol{I}_{t+1}(\boldsymbol{\psi}_{t+1})$. Adding and
subtracting $\boldsymbol{I}_k(\boldsymbol{\psi}_t)$ inside the bracket of the resulting
difference and factoring out the leading Jacobian $\boldsymbol{J}_{t+1}$,
\begin{equation}
    \boldsymbol{S}_{t+1}^\ast - \tilde{\boldsymbol{S}}_{t+1}
    \;=\; \boldsymbol{J}_{t+1}\!\sum_{k=0}^{t} \left(\prod_{j=k+1}^{t} \boldsymbol{J}_j\right)\!
    \bigl[\boldsymbol{I}_k(\boldsymbol{\psi}_{t+1}) - \boldsymbol{I}_k(\boldsymbol{\psi}_t)\bigr]
    \;+\; \boldsymbol{J}_{t+1}\bigl(\boldsymbol{S}_t^\ast - \tilde{\boldsymbol{S}}_t\bigr).
\end{equation}
Taking norms, applying (A1) to each product of Jacobians, applying (A2)
and (A3) to the freshly injected term, and bounding the geometric sum by
$1/(1-\rho)$,
\begin{equation}
    E_{t+1} \;\le\; \rho \cdot \frac{L_I\, \eta\, G}{1 - \rho}
    \;+\; \rho\, E_t.
    \label{eq:err_recur}
\end{equation}
This is a first-order linear recurrence with contraction factor
$\rho < 1$ and constant driving term. The transient $\rho^t E_0$
vanishes and the geometric series converges, yielding the steady-state
bound
\begin{equation}
    E_\infty \;\le\; \frac{\rho\, L_I\, G\, \eta}{(1 - \rho)^2}.
    \label{eq:err_bound}
\end{equation}

\paragraph{Periodic updates.}
If $\boldsymbol{\psi}$ is held fixed for $m$ steps between updates, the staleness
injected at each update decays by $\rho^m$ before the next, sharpening
the bound to
\begin{equation}
    E_\infty \;\le\;
    \frac{\rho\, L_I\, G\, \eta}{(1 - \rho)(1 - \rho^m)}.
\end{equation}
Since $1 - \rho^m \ge 1 - \rho$ for $m \ge 1$, periodic updates strictly
reduce the steady-state staleness.

\subsection{A first-order Taylor-trace correction}
\label{sec:staleness-correction}

The leading-order term in the error recurrence~\eqref{eq:err_recur} is
the freshly injected staleness
$\boldsymbol{I}_k(\boldsymbol{\psi}_{t+1}) - \boldsymbol{I}_k(\boldsymbol{\psi}_t)$. A first-order Taylor expansion in
$\boldsymbol{\psi}$ gives
\begin{equation}
    \boldsymbol{I}_k(\boldsymbol{\psi}_t) \;\approx\; \boldsymbol{I}_k(\boldsymbol{\psi}_{t-1})
    \;+\; \left.\frac{\partial \boldsymbol{I}_k}{\partial \boldsymbol{\psi}}\right|_{\boldsymbol{\psi}_{t-1}}\!\!
    \Delta\boldsymbol{\psi}_{t-1},
\end{equation}
where $\Delta\boldsymbol{\psi}_{t-1} = \boldsymbol{\psi}_t - \boldsymbol{\psi}_{t-1}$. Substituting into
$\boldsymbol{S}_t^\ast$ and grouping the second-order factors into an auxiliary
accumulator
\begin{equation}
    \boldsymbol{\Omega}_t \;=\; \boldsymbol{J}_t\, \boldsymbol{\Omega}_{t-1}
    \;+\; \left.\frac{\partial \boldsymbol{I}_t}{\partial \boldsymbol{\psi}}\right|_{\boldsymbol{\psi}_t}
\end{equation}
yields a corrected update
\begin{equation}
    \boldsymbol{S}_t \;=\; \boldsymbol{J}_t\,\bigl(\boldsymbol{S}_{t-1} + \boldsymbol{\Omega}_{t-1}\, \Delta\boldsymbol{\psi}_{t-1}\bigr)
    \;+\; \boldsymbol{I}_t(\boldsymbol{\psi}_t).
\end{equation}
The additive term $\boldsymbol{J}_t\, \boldsymbol{\Omega}_{t-1}\, \Delta\boldsymbol{\psi}_{t-1}$ neutralizes
the leading-order staleness injection, reducing the residual error from
$\mathcal{O}(\|\Delta\boldsymbol{\psi}\|)$ to $\mathcal{O}(\|\Delta\boldsymbol{\psi}\|^2)$.

Storing $\boldsymbol{\Omega}_t$ exactly is intractable: since $\boldsymbol{I}_t$ has shape
$|h| \times p$, $\boldsymbol{\Omega}_t$ is a rank-three tensor of shape
$|h| \times p \times p$ requiring $\mathcal{O}(|h|\, p^2)$ memory. A
parameter-wise diagonal approximation,
\begin{equation}
    \boldsymbol{\omega}_t \;=\; \boldsymbol{J}_t\, \boldsymbol{\omega}_{t-1}
    \;+\; \mathrm{diag}_{\boldsymbol{\psi}}\!\left(\frac{\partial \boldsymbol{I}_t}{\partial \boldsymbol{\psi}}\right),
    \quad
    \boldsymbol{S}_t \;=\; \boldsymbol{J}_t\,\bigl(\boldsymbol{S}_{t-1} + \boldsymbol{\omega}_{t-1} \odot \Delta\boldsymbol{\psi}_{t-1}\bigr)
    \;+\; \boldsymbol{I}_t(\boldsymbol{\psi}_t),
    \label{eq:taylor-diag}
\end{equation}
recovers the original space complexity of the \gls{rtrl} sensitivity,
where $\odot$ denotes the elementwise product along the parameter axis.

\makeatletter
\if@accepted

\newpage
\subsubsection*{Acknowledgments}
\label{sec:ack}
This work was funded by the Deutsche Forschungsgemeinschaft (DFG, German Research Foundation) under Germany's Excellence Strategy (EXC-3057/1 ``Reasonable Artificial Intelligence'', Project No. 533677015).
\\
Noah Farr is supported by the Konrad Zuse School of Excellence in Learning and Intelligent Systems (\textbf{\href{https://eliza.school/}{ELIZA}}) through the DAAD programme Konrad Zuse Schools of Excellence in Artificial Intelligence, sponsored by the Federal Ministry of Education and Research.
\fi
\makeatother


\bibliography{main}
\bibliographystyle{rlj}

\end{document}